\documentclass{article}
\usepackage{ijcai22}

\usepackage{times}
\usepackage{soul}
\usepackage{url}
\usepackage{subfig}
\usepackage[hidelinks]{hyperref}
\usepackage[utf8]{inputenc}
\usepackage{graphicx}
\usepackage{amsmath}
\usepackage{amsthm}
\usepackage{amssymb}
\usepackage{booktabs}
\usepackage{algorithm}
\usepackage{algorithmic}
\usepackage{siunitx,booktabs}
\usepackage{enumitem}
\usepackage{parskip}

\newcommand{\ie}{\textit{i}.\textit{e}.}

\newcommand{\vpara}[1]{\vspace{0.05in}\noindent\textbf{#1 }}

\title{A Novel Architecture Slimming Method for Network Pruning and Knowledge Distillation}

\author{
Dongqi Wang$^1$\and
Shengyu Zhang$^2$\and
Zhipeng Di$^1$ \and
Xin Lin$^1$\and
Weihua Zhou$^1$\thanks{Corresponding author: larryzhou@
zju.edu.cn} \and 
Fei Wu$^2$\\
\affiliations
$^1$School of Management, Zhejiang University\\
$^2$College of Computer Science and Technology, Zhejiang University\\
\emails
\{lightningwdq,d1169028312\}@gmail.com,
\{sy\_zhang,linx612,larryzhou,wufei\}@zju.edu.cn
}

\begin{document}

\maketitle

\begin{abstract}

Network pruning and knowledge distillation are two widely-known model compression methods that efficiently reduce computation cost and model size. A common problem in both pruning and distillation is to determine compressed architecture, \ie, the \textbf{exact} number of filters per layer and layer configuration, in order to preserve most of the original model capacity.  In spite of the great advances in existing works, the determination of an excellent architecture still requires human interference or tremendous experimentations. In this paper, we propose an architecture slimming method that automates the layer configuration process. We start from the perspective that the capacity of the over-parameterized model can be largely preserved by finding the minimum number of filters preserving the maximum parameter variance per layer, resulting in a thin architecture. We formulate the determination of compressed architecture as a one-step orthogonal linear transformation, and integrate principle component analysis (PCA), where the variances of filters in the first several projections are maximized. We demonstrate the rationality of our analysis and the effectiveness of the proposed method through extensive experiments. In particular, we show that under the same overall compression rate, the compressed architecture determined by our method shows significant performance gain over baselines after pruning and distillation. Surprisingly, we find that the resulting layer-wise compression rates correspond to the layer sensitivities found by existing works through tremendous experimentations.

\end{abstract}

\section{Introduction}


Neural network pruning and knowledge distillation have become the main force of model compression methods with the common goal to reduce high demand on the memory footprint and computing resource while maintaining accuracy by learning compresssed networks~\cite{yao2021edge}. 
As for pruning, weight pruning approaches preserve model generalization performance by a pruned sparse structure, which however is limited on general-purpose hardware or BLAS libraries. In contrast, filter pruning works at the filter level by removing unnecessary filters, resulting in a dense structure that can be deployed on edge computing devices.
Not surprisingly, the pruned architecture is found of significant influence on performance\cite{liu2018rethinking}. However, most mainstream cutting-edge methods are based on expert-designed handcraft rules, requiring massive human interference. A common criterion is the ratio of channels to prune in each layer. 
Knowledge distillation \cite{Hinton_Vinyals_Dean_2015} is another well-known model compression technique, which distills knowledge from a cumbersome teacher model into a light-weight student model without losing much model capacity.
Similarly to network pruning, the student architecture is found as a contributing factor to its learning ability \cite{Liu_Jia_Tan_Vemulapalli_Zhu_Green_Wang_2020}.
However, many existing works directly adopt an expert-generated light-weight architecture (such as MobileNets \cite{Howard_Zhu_Chen_Kalenichenko_Wang_Weyand_Andreetto_Adam_2017}) as the student \cite{Yun_Park_Lee_Shin_2020} without considering whether it can preserve the model capability of the teacher architecture, so as to easily cause the model capacity gap.\\
To solve this problem, some automatic pruning methods and automatic student architecture search methods has been put forward\cite{lin2020channel,he2018amc}, which can search and find network without hand-craft rules. Among them,for network pruning,the heuristic-based and reinforcement learning-based pruning methods have become the mainstream methods.Their core idea is to perform the optimal layer-wise search in all possible cases on a certain fine-grained basis and then find the best one as the result. This eliminates the need for a lot of manual time and effort to configure the network.However, these methods have shortcomings in convergence \cite{Li_Gu_Zhang_Gool_Timofte_2020}. The convergence uncertainty that corresponds directly to the fine-grained settings makes the results of these methods not guaranteed.There is a trade-off bewtween accuracy and computing resource. In addition, for knowledge distillation, existing student determination methods involve replacing teachers' expensive operations with cheaper ones \cite{Crowley_Gray_Storkey_2018}, and neural architecture search (NAS) \cite{Liu_Jia_Tan_Vemulapalli_Zhu_Green_Wang_2020}. However, we might not always find cheaper operations for a given teacher architecture. Search-based methods typically require tremendous experimentations through multi-turn student generation and evaluation, which might take up to 5 days on 200 TPUs \cite{Liu_Jia_Tan_Vemulapalli_Zhu_Green_Wang_2020}.
Therefore, for both network pruning and knowledge distillation, we look for a automatic layer-wise architecture determination method that is efficient, \ie, without multi-turn search, and effective, \ie, with theoretical guarantee on the preserved model capability.
\begin{figure}
\centering
\subfloat[cumulative variance contribution in each layer]{\includegraphics[scale=0.04]{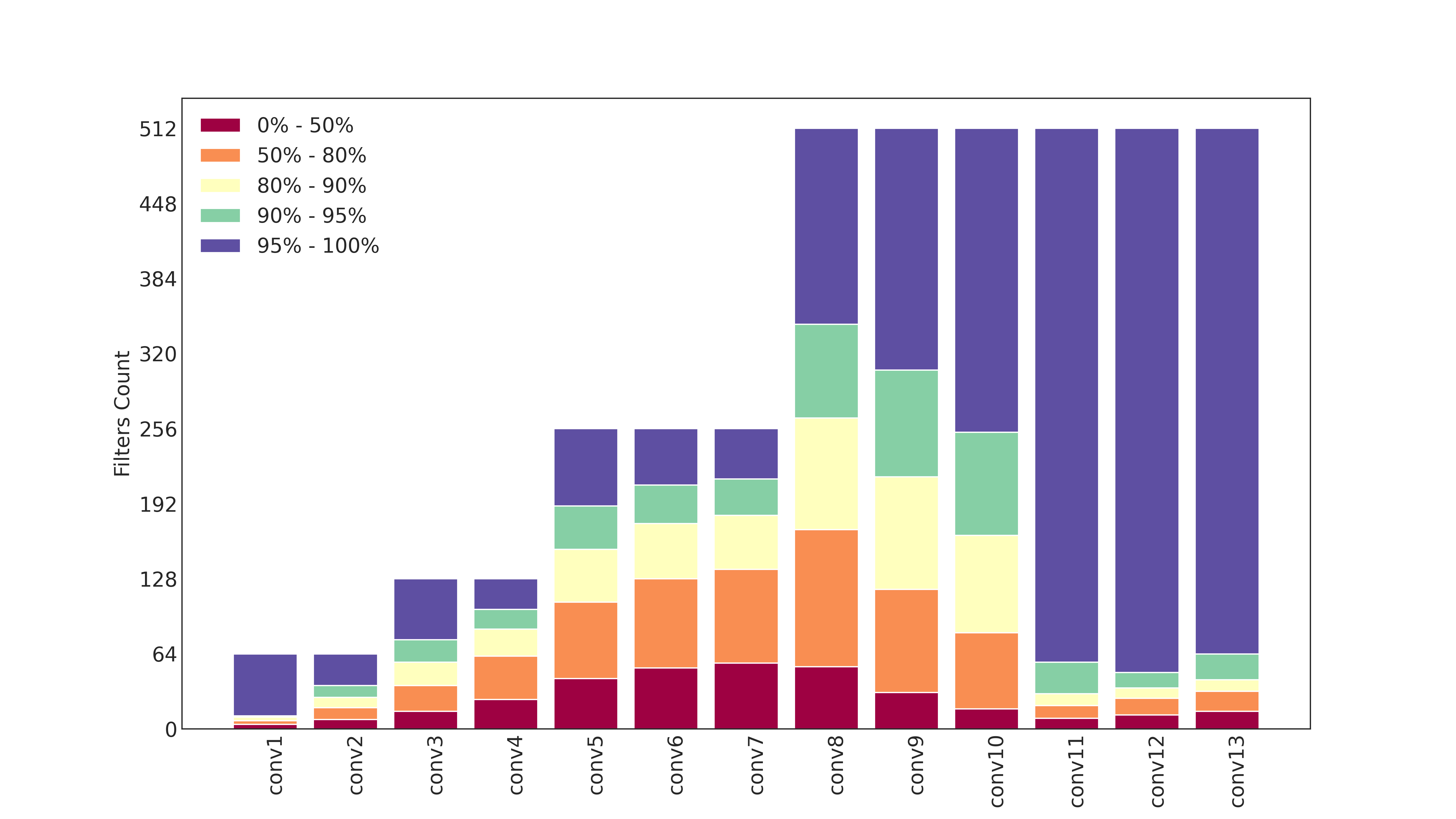}\label{fig:fig1A}}\\
\subfloat[cumulative variance contribution rate in conv10]{\includegraphics[scale=0.04]{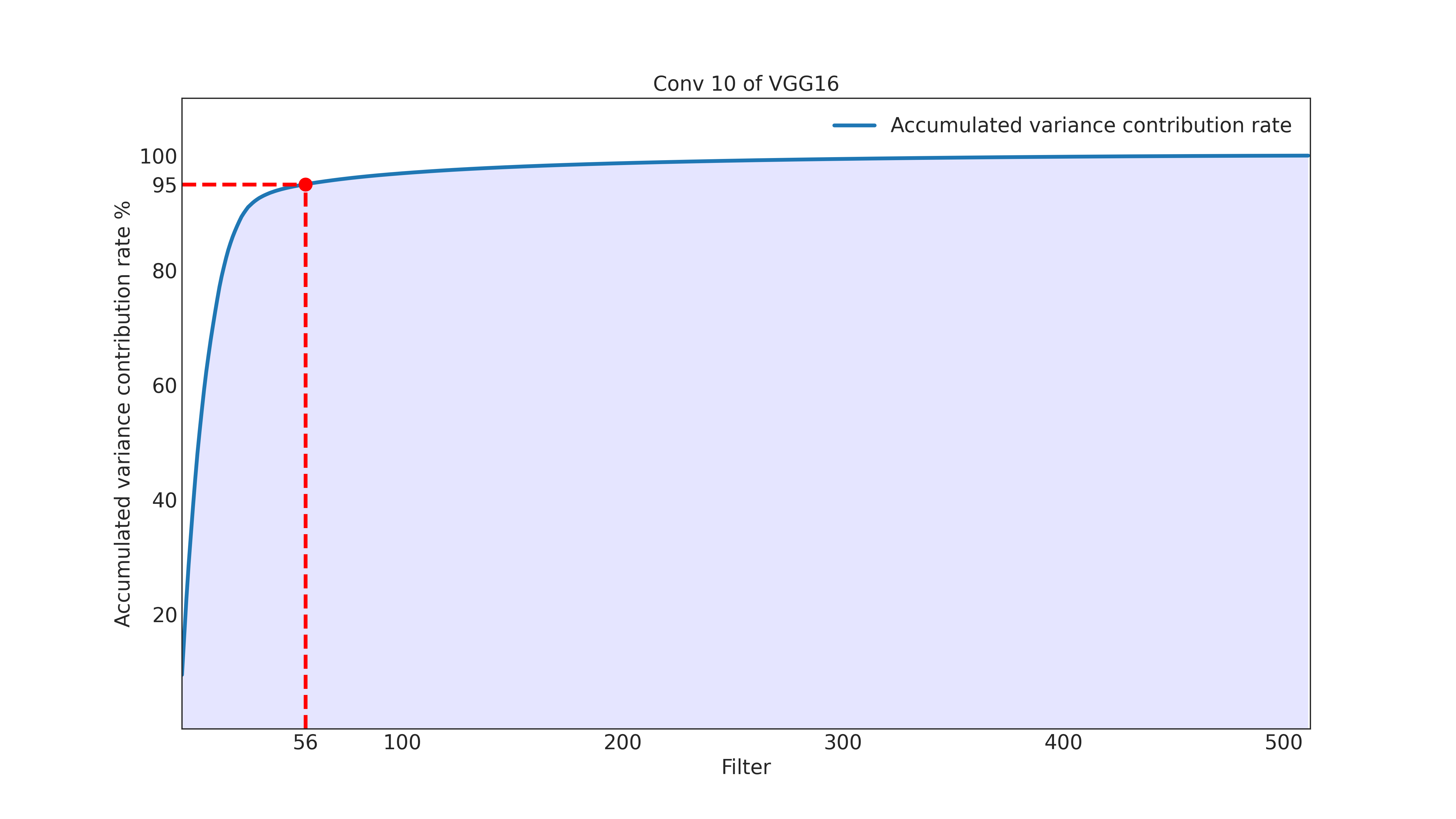}\label{fig:fig1B}}
\caption{VGG trained on Cifar10 dataset, and do a one-step orthogonal linear transformation via PCA. It known as an over-parameterized model, thus the information variance in each layer can be characterized by less filters.}
\end{figure}\\
To bridge the gap, we propose an automatic adaptive architecture slimming method for network pruning and knowledge distillation, especially for CNN-based neural networks. For an over-parameterized model, we propose to conduct layer-wise parameter configuration, and find the minimum number of filters to preserve the maximum model capacity per layer. Inspired by information theory where varainace of information is a metric to information\cite{arabie1973multidimensional}, we link model capacity to parameter diversity \cite{Ayinde_Inanc_Zurada_2019}, and more specifically, filter variance, which is measurable. Then, the problem has been changed to find the minimum number of filters that could preserve the most filter variance per layer.
We integrate principle component analysis, and conduct orthogonal linear transformations to existing filters, where we expect the first several dimensions after projection can preserve the most filter variance. As shown in Figure 1(a), for each layer in VGG trained on Cifar10 dataset, after probjection, we can preserve the 95\% varaince by obviously less filters(the part of the bar chart except blue part), to be more specificly, as shown in (b),the cumulative variance varies in conv10 with the number of filters, we can just use 56 filters(about 10\%) to preserve the information represented by the original 512 filters, and cumulative filter variance increases sharply for the first several projected dimensions and quite smoothly for the remaining ones, which serves as an empirical evidence for the rationality of our method.\\
We conduct extensive experiments using standard network pruning and knowledge distillation evaluation protocols. We show that under a given overall compression rate, the thin architecture found by our method leads to significantly better performance in both pruning and distillation compared to existing baselines and across various backbone architectures. Remarkably, we find that the layer-wise compression rates found by our method approximate the layer sensitivities found by existing works \cite{Li_Kadav_Durdanovic_Samet_Graf_2017a} through expensive experimentations.
To summarize, this paper makes the following key contributions:
\begin{itemize}
    \item We emprically show that layer-wise filter variance is a feasible and effective indicator for pruned/student architecture determination.
    \item We propose a model-agnostic architecture slimming method based on principle component analysis, where the filter variance preserved by the first several dimensions is maximized after othogonal probjection. 
    \item Extensive experiments on network pruning and knowledge distillation demonstrates the rationality of our analysis and the effectiveness of our method.
\end{itemize}

\section{Related Works}

\textbf{Knowledge Distilling}. Knowledge distillation \cite{Hinton_Vinyals_Dean_2015} learns a compressed student model from a cumbersome teacher that is mostly over-parameterized (\textit{e.g.}, pre-trained large models~\cite{zhang2020devlbert,yao2022musicrepresentation}). The knowledge distilled exhibits different kinds, ranging from response-based knowledge \cite{Hinton_Vinyals_Dean_2015}, feature-based knowledge \cite{Romero_Ballas_Kahou_Chassang_Gatta_Bengio_2015}, and relation-based knowledge \cite{Yim_Joo_Bae_Kim_2017}. Modern distillation technique includes offline distillation \cite{Hinton_Vinyals_Dean_2015} where the teacher is typically pre-trained and fixed, online distillation where the teacher is simutaneously updated, and self-distillation \cite{Zhang_Song_Gao_Chen_Bao_Ma_2019} where the teacher and the student share the same parameters and the architecture. Recently, besides the knowledge forms and the distllation techniques, the student architecture has been found to largely contribute to the distillation effectiveness \cite{Liu_Jia_Tan_Vemulapalli_Zhu_Green_Wang_2020}. Currently, the determination of the student architecture is mostly human-interfered by mostly choosing a simplified version of the teacher network \cite{Wang_Zhao_Li_Tan_2018} or another architecture with efficient operations. More recently, there are a few works that exploit neural architecture search \cite{Eyono_Carlucci_Esperana_Ru_Torr_2021}, which, however, requires tremendous experimentations and takes up to 5 days on 200 TPUs. In this paper, we propose an automatic and adaptive technique that takes only one-step linear probjection to find the optimal filter combination to mostly preserve the original capacity for each layer in the teacher model, resulting in an effective student model for distillation.\\
\textbf{Network pruning}.
Network pruning, aim to cut off the unimportant parts or rewrite the parameter by pruning in the form of constrained optimization. Filter pruning removes the entire redundant filters directly according to certain metrics, not only significantly reduces storage usage, but also decreases the computational cost in online inference, which can be well supported by general-purpose hardware and high-efficiency Basic Linear Algebra Subprograms (BLAS) library. Existed filter pruning works mainly based on some criterions,\cite{Li_Kadav_Durdanovic_Samet_Graf_2017a,he2019filter}, the above methods have good performance, but they takes a lot of time and computing resources to find a suitable structure in a certain classification task.
With the development of AutoML, more automatic methods have been proposed on network pruning. It selects the best condidates from all possiblities of subnets.It mainly consists of the heuristic-based and reinforcement learning-based pruning methods. \cite{he2018amc} leverages reinforcement learning to provide the model compression policy with a good compression performance. As far as heuristic-based pruning, \cite{lin2020channel} uses artificial bee colony algorithm (ABCPruner) to efficiently find pruned structure. As it’s mentioned in introduction, most of the automatic pruning methods are faced with the concern of convergence. Therefore, a lot of calculation and time costs are inevitable.

\section{The Automatic Adaptive Architecture Slimming Method (3AS)} 

\subsection{Overview}
\begin{figure}
\centering
\subfloat[filter pruning process]{\includegraphics[scale=0.4]{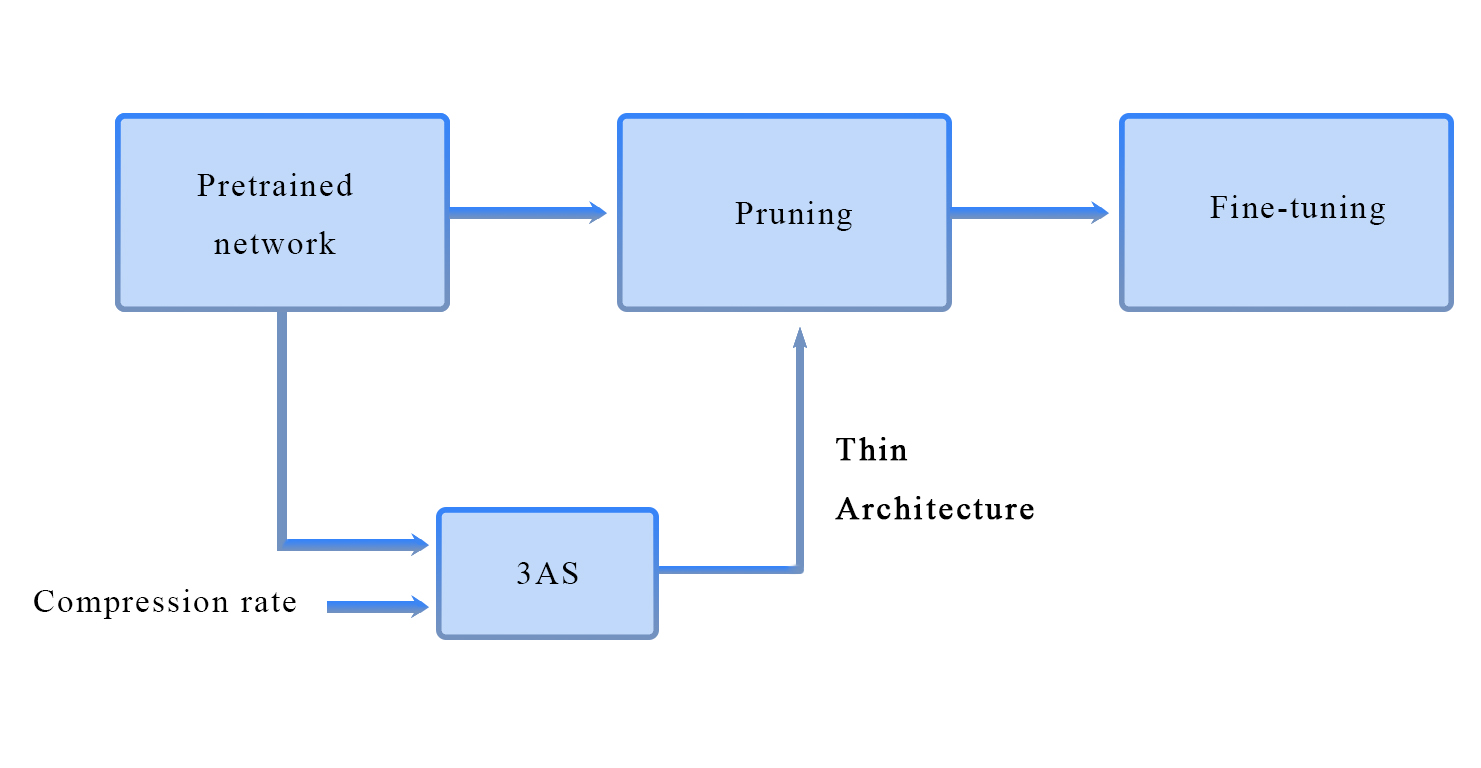}\label{fig:fig2A}}\\
\quad
\subfloat[knowledge distillation process]{\includegraphics[scale=0.4]{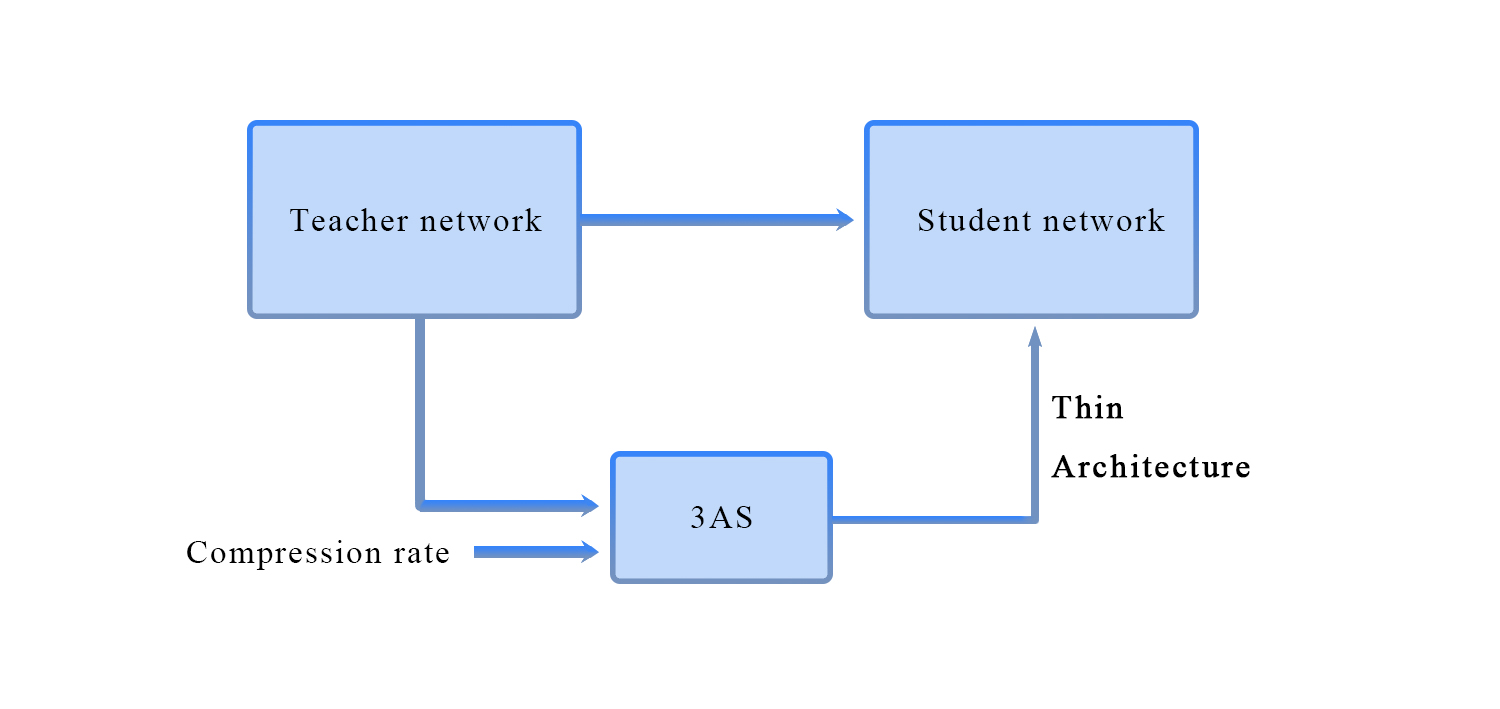}\label{fig:fig2B}}
\caption{3AS automates the 3-stage pruning process and knowledge distillation process}
\end{figure}
In 3AS, the filters in each layer will be squeezed into a 2-dimension tensor where the two dimensions represent the number of filters and the product of the number of channels and the kernel size, respectively. Since the main idea of 3AS to preserve model capacity by mamizing variance of remaining filters, we firstly give an illustration on why variance is a good indicator of information in Section \ref{sec:pre}. We give a proof of the information lossless property of orthogonal transformation in Section \ref{sec:ortho}. Upon the guarantee, an orthogonal linear transformation with threshold $\alpha$ will maximize the parameter variance in the first several projected dimensions. The percentage of variance preserved will be used as the guidance for determining the number of filters of this layer in the compressed thin structure. This thin structure will be served as the structure prior for a particular pruning method or as the student model in knowledge distillation. As shown in figure2(a), for network pruning the proposed method can adaptively determine the pruned architecture based on compression demand and pretrained model,it automates the common 3-stage pruning process. In figure2(b), for knowledge distilling, the proposed method provides an approprite architecture as the student network based on teacher network and compression demand. We give a formal defintion of the architecture optimization model in Section \ref{sec:model}, and elaborate the detailed process of the 3AS algorithm in Section \ref{sec:3AS}.\\
\subsection{Preliminaries} \label{sec:pre}
We will formally introduce the symbols and annotations in this subsection. Given a pre-trained network $W$, which contains the convolutional layers set $L$ 
where $\forall l \in \left\{1,2,\ldots,\left|L \right|\right\}$.
Each layer consists of a parameterized filters set $\boldsymbol{F}_l$
where $\forall j\in \left\{1,2,\cdots,|\boldsymbol{F}_{l}|\right\}$. Denote the $i^{th}$ parameter in filter $j$ as $w^{\left(l\right)}_{i,j} \in \mathbb{R}^{\left|\boldsymbol{F}_{l-1} \right|\times k_{l}\times k_{l}}$, 
with kernal size $k_{l}\times k_{l}$. 
We regrad the parameters vector $\boldsymbol{w}^{\left(\boldsymbol{F}_l\right)}_{i}$
as a point in orthogonal space $\boldsymbol{F}_l$ in Figure 3,
with the form of
$\left[w^{\left (l\right)}_{i,1},w^{\left (l\right)}_{i,2},\ldots,w^{\left (l\right)}_{i,\left|\boldsymbol{F}_l\right|}\right]^{T}, \forall i\in\left\{ 1,2,\ldots,k_l\times k_l \right\}$.
Without loss of generality, 
we assume the $\boldsymbol{w}^{\left(\boldsymbol{F}_l\right)}_{i}$
obeys a multivariate Gaussian distribution as
$\boldsymbol{N}_{\left (\boldsymbol{F}_l\right)}$
, where
\centerline{
$\boldsymbol{w}^{\left(\boldsymbol{F}_l\right)}_{i}
\sim \boldsymbol{N}_{\left(\boldsymbol{F}_l\right)} \left(\boldsymbol{\mu}^{\left(\boldsymbol{F}_l\right)},\boldsymbol{\varSigma}^{\left(\boldsymbol{F}_l\right)}\right)$
}
The trace of variance-covariance matrix measures a distance, namely the information gap, between each point and their geometric center. Thus, the measurement of total amount of information $\Psi
_{\boldsymbol{F}_{l}}$, which stored at a group of filters, can be denoted as follow.
\\
\centerline
{$
\Psi_{\boldsymbol{F}_l}=\frac{1}{k^2_{l}}\left\| \boldsymbol{\mathrm{tr}}\left(\boldsymbol{\varSigma}^{\left(\boldsymbol{F}_l\right)}\right) \right\|_1,\forall l \in L
$}
In probability theory and information theory, the variation of information or shared information distance is a measure of the distance between two clusterings (partitions of elements).
\cite{arabie1973multidimensional}
It is a true metric when filters are the information carriers in each layer, and we extend our analysis as bellow under this view. 

\begin{figure}
\includegraphics[scale=0.4]{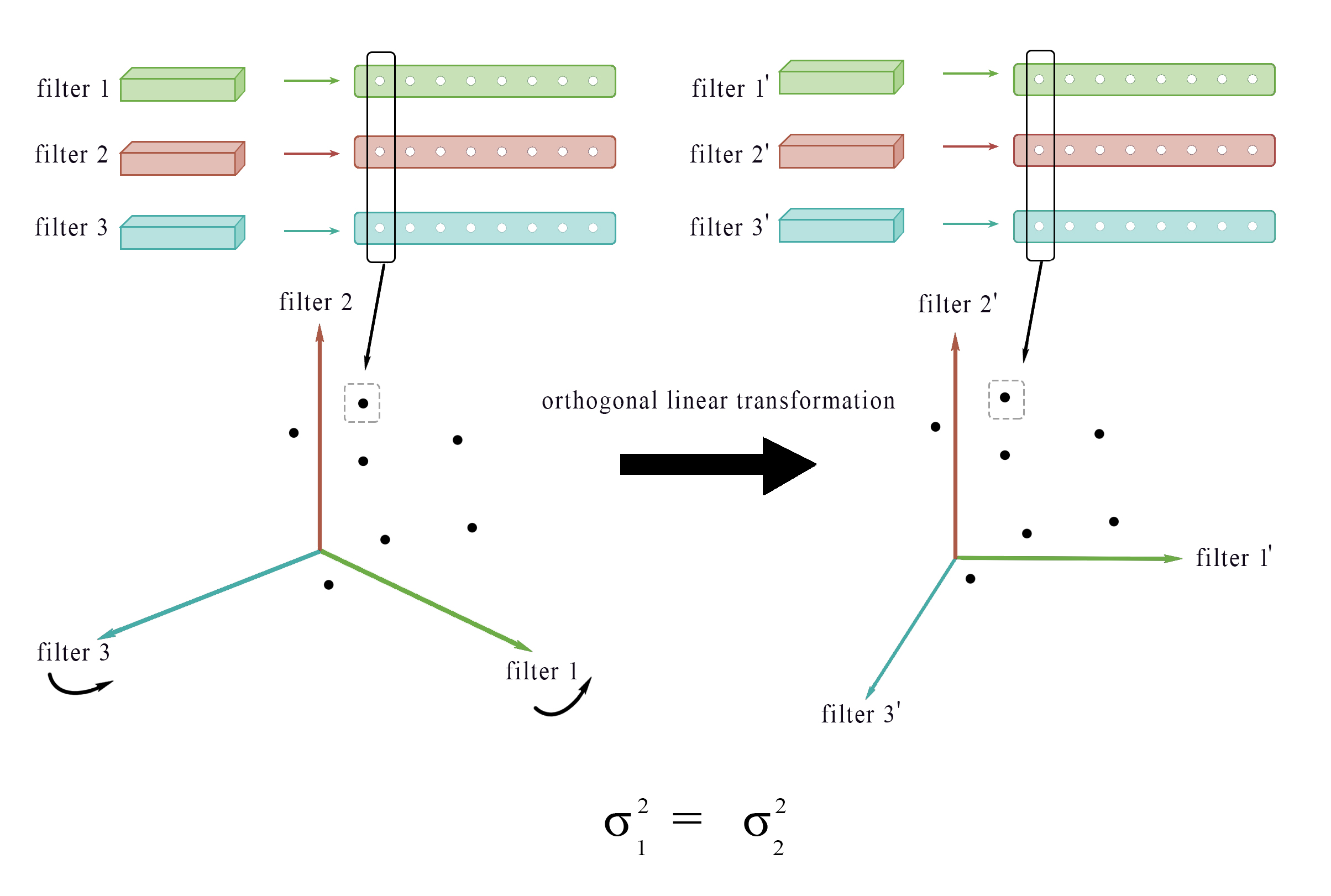}
\caption{One-step orthogonal transformation in 3AS: Squeezed parameters space in 2 dimensions, let each filter as an axis and rotate the whole coordinate system in invariant variance.}
\end{figure}

\subsection{An Information Lossless Process: The Same Dimensional Orthogonal Transformation} \label{sec:ortho}
Here, we anaylze the amount of information $\Psi_{\boldsymbol{F}_l}$ will keep constant after any $|\boldsymbol{F}_l|$-dimensional orthogonal transformation. Firstly, we introduce the \textbf{Lemma 1} to show the invariance of distribution $\boldsymbol{N}_{\left (\boldsymbol{F}_l\right)}$ through an orthogonal reflection.

\textbf{Lemma 1}: Under any orthogonal transformation, the independence and homoscedasticity of multivariate normal distribution $\boldsymbol{N}$ remain unchanged.

$Proof$:
The orthogonal matrix $\boldsymbol{C}_{n\times n}$ triggers the reflection from $\boldsymbol{N}\left(\boldsymbol{\mu},\boldsymbol{\varSigma}\right)$ 
to 
$\boldsymbol{N}\left(\boldsymbol{C\mu},\boldsymbol{C\varSigma{C^T}}\right)$.
According to the property of $\boldsymbol{C}$:
$\sum_{j=1}^n{c_{ij}^{2}=1}$, $\sum_{j=1}^{n} c_{ji}c_{ki}=0$, $\forall k\in \{ 1,2,\ldots,n \} \backslash\{ j \}$.
Easily to derive $\boldsymbol{C\varSigma{C^T}}=\boldsymbol{\varSigma}$. Proofed.

Then, we suppose a new orthogonal space $\boldsymbol{S}_{l}$. The invariance information sharing: $\boldsymbol{F}_{l}\mapsto\boldsymbol{S}_{l}$ is verified in \textbf{Corollary 1}. 

\textbf{Corollary 1}:
$
\Psi_{\boldsymbol{F}_l}=\Psi_{\boldsymbol{S}_l}
$ 
if 
$\left| \boldsymbol{S}_{l} \right|=\left| \boldsymbol{F}_{l} \right|$.

$Proof$: Denote the distance of  point $\boldsymbol{w}^{\left(\boldsymbol{F}_l\right)}_{i}$ to geometric center as $\sigma^2_{l,i}$. It can be decompose to each axis in $\boldsymbol{S}_{l}$ with $\sigma^2_{l,i,j}$. 
Accroding to \textbf{Lemma 1} when $\left| \boldsymbol{S}_{l} \right|=\left| \boldsymbol{F}_{l} \right|$, there has:\\
\centerline
{
$\sigma^2_{l,i}
=\sum_{m=1}^{\left| S_l \right|}{\sigma _{l,i,m}^{2}}$
and 
$\sigma^2_{l,i}
=\sum_{j=1}^{\left| F_l \right|}{\sigma _{l,i,j}^{2}}$
}
By the defination of variance-covariance matrix, we have $\sum_{i=1}^{k_l\times k_l}{\sigma_{l,i}^{2}}=\left\| \boldsymbol{\mathrm{tr}}\left(\boldsymbol{\varSigma}_{\left(\boldsymbol{F}_l\right)}\right) \right\|_1$. 
Thus, 
$\Psi_{\boldsymbol{F}_l}=\Psi_{\boldsymbol{S}_l}$.
Proofed.

The \textbf{Corollary 1} indicates that a completed parameters set stored in specific cell in a group of filters will never loss any information through any different orthogonal transformation. Although $\boldsymbol{w}^{\left(\boldsymbol{F}_l\right)}$ in space $\boldsymbol{F}_l$ and $\boldsymbol{S}_l$ have same variance, the variance accumulate contribution at each axises is not the same, i.e. ${\sigma_{l,i,m}^{2}}\neq{\sigma_{l,i,j}^{2}}, m=j$. 
This phenomenon inspired our work in the next subsection.

\subsection{The Architecture Optimization Model under Specific Slimming Target} \label{sec:model}
In this subsection, we formulate a mathematical model to determine the optimal filters number in specific layer. Given a group of orthogonal space $\boldsymbol{S}_\Theta$ including $\boldsymbol{S}_l$ , we denote a binary variable ${x}^{(\boldsymbol{S}_l)}_{m}$ for selecting representative dimension:\\
\centerline{
$x_{m}^{\left(\boldsymbol{S}_l \right)}=\begin{cases}
	1, \mathrm{reserve\,\,the \,\,dimension}\,\,m\\
	0, \mathrm{otherwise}\\
\end{cases}$
}
The goal of architecture optimization is to find a feasible $\boldsymbol{S}_l$ in $\boldsymbol{S}_\Theta$ for reserving a smaller number of dimensions, meanwhile the cumulative variance contribution rate in this reserved space achieves a slimming target $\delta$. Therefore we formulate the mathematical model when $0\le\delta\le1$ and $
\forall i \in \{1,2,\ldots,k^{2}_{l}\},
\forall l \in L
$:
\begin{equation}
    \underset{\boldsymbol{S}_l\in \boldsymbol{S}_{\Theta}}{arg}\min \sum_{m=1}^{\left|\boldsymbol{S}_l \right|}{x_{m}^{\left(\boldsymbol{S}_l \right)}},
\end{equation}
\begin{equation}
    s.t.
    \sum_{m=1}^{\left|\boldsymbol{S}_l \right|}{x_{m}^{\left(\boldsymbol{S}_l \right)}\sigma _{l,i,m}^{2}}\ge\delta    \Psi_{\boldsymbol{F}_l},
\end{equation}
For $|\boldsymbol{S}_l|=|\boldsymbol{F}_l|$, after deleting the worst dimension which has the lowest variance in $\boldsymbol{S}_l$, we denote the reserved space in $\boldsymbol{S}^{({|\boldsymbol{F}_l|}-1)}_{l}$. Then, we can construct a sequencial space $\{\boldsymbol{S}^{({|\boldsymbol{F}_l|}-1)}_l,\boldsymbol{S}^{({|\boldsymbol{F}_l|}-2)}_l,\ldots,\boldsymbol{S}^{(1)}_l\}$. 
It is easily to solve the one-dimensional model and output the solution $\boldsymbol{S}^{(1)*}_l$. 
The $\boldsymbol{S}^{(2)*}_l$ 
is also availabe to solve as we know $\boldsymbol{S}^{(1)*}_l$. 
By parity of reasoning, The optimal solution $\boldsymbol{S}^{(n)*}_l$ 
is stopped finding when the condition of equation (2) is satisfed. Thus the optimal solution of filters archetecture in layer $l$ can be outputted as $n^{*}_l$, within a one-step orthogonal mapping $\boldsymbol{F}_l\mapsto\boldsymbol{S}^{(n)*}_l\cap\boldsymbol{S}_l, \forall\boldsymbol{S}_l\in \boldsymbol{S}_{\Theta}$. 

\subsection{The 3AS Algorithm} \label{sec:3AS}
The solving process we mentioned above has a similar perspective with the principal component analysis(PCA) framework. The optimal guarantee for our solution is consistent with the \textbf{Property 1} in PCA framework, which also can be concluded with \textbf{Lemma 1} and \textbf{Corollary 1}.\\
\textbf{Property 1}:
For any integer $n$, $1\le n\le |\boldsymbol{F}_l|$, consider the orthogonal linear transformation $y=\boldsymbol{C}^{T}\boldsymbol{w}^{(\boldsymbol{F}_l)}$, where $y$ is a n-element vector and $\boldsymbol{C}$ is a $n\times|\boldsymbol{F}_l|$ matrix.
Let $\boldsymbol{\varSigma}_{y}=\boldsymbol{C}_{T}\boldsymbol{\varSigma}\boldsymbol{C}$ be the variance-covariance matrix for $y$. Then $\boldsymbol{\mathrm{tr}}(\boldsymbol{\varSigma}_{y})$ is maximized by taking $\boldsymbol{C}=\boldsymbol{A}_n$, where $\boldsymbol{A}_n$ consists of the first $n$ columns of $\boldsymbol{A}$.

The steps of 3AS algorithm are proposed as follow:\\
\textbf{Step 1}:
For a convolutional layer $l$, We squeeze its dimension from $\boldsymbol{w}_{l}\in \mathbb{R}^{|\boldsymbol{F}_{l}|\times |\boldsymbol{F}_{l-1}|\times k_{l}\times k_{l}}$ to 
$\boldsymbol{\varpi}_{l}\in \mathbb{R}^{\boldsymbol{F}_{l}\times (|\boldsymbol{F}_{l-1}|\times k^2_{l})}$, 
namely we have squeezed the dimensions of each layer from four dimensions to two dimensions. Calculate its transposition as $\boldsymbol{\varpi}^{T}_{l}\in \mathbb{R}^{(|\boldsymbol{F}_{l-1}|\times k^2_{l})\times\boldsymbol{F}_{l}}$.\\
\textbf{Step 2}:
After $\boldsymbol{\varpi}_{l}$ normalizing by column, we construct the covariance matrix $\boldsymbol{\varSigma}_{l}=\boldsymbol{\varpi}_{l}\boldsymbol{\varpi}_{l}^T$ and do singular value decomposition(SVD) on $\boldsymbol{\varSigma}$ as 
\begin{equation}
    \boldsymbol{\varSigma}=\boldsymbol{U}\boldsymbol{\Omega}\boldsymbol{U}^{T}=\sum_{i=1}^{|\boldsymbol{F}_{l}|}\lambda_{i}\boldsymbol{u}_{j}\boldsymbol{u}_{j}^T
    \forall l \in L
\end{equation}
Where
$\boldsymbol{U}=(\boldsymbol{u}_{1},\boldsymbol{u}_{2},\dots,\boldsymbol{u}_{|\boldsymbol{F}_{l}|})$ 
is an orthonormal matrix and $\boldsymbol{\Omega}=diag(\lambda_{1},\lambda_{2},\dots,\lambda_{|\boldsymbol{F}_{l}|})$ 
with
$\lambda_{1}\geq\lambda_{2}\geq\dots\geq\lambda_{|F_{l}|}$is a diagonal matrix.
$\boldsymbol{\lambda}$ is the eigenvalue of $\Omega$, to evaluate the degree of information integrity.\\
\textbf{Step 3}:
Calculate the cumulative contribution rate:
\begin{equation}
    \alpha_{n_{l}}=\frac{\sum_{i=1}^{n_{l}}\lambda_{i}}{\sum_{i=1}^{|\boldsymbol{F}_{l}|}\lambda_{i}}
    \forall l \in L
\end{equation}
Where $n_{l}$ is the principal component retention quantity we tend to preserve.\\
\textbf{Step 4}:
Confirm the value of $n_{i}$ when the cumulative contribution rate achieve the target $\delta$: 
\begin{equation}
    n^{*}_{l}=\arg\min_{n_{l}}\{\alpha_{n_{l}}|\alpha_{n}\ge\delta\},
    \forall l \in L 
\end{equation}
To summarize, the detail of 3AS lists in \textbf{Algorithm 1}.

\begin{algorithm}[tb]
\caption{Description of 3AS}
\label{alg:algorithm}

\textbf{Input}: $W$: pre-trained network \\ 
                $\delta$: The threshold value of 3AS \\
\textbf{Initialize}: $|\boldsymbol{F}_{l}|$: Number of filters at layer $l$ \\ 
                $k_{l}$: The size of filter in layer $l$  \\ 
\begin{algorithmic}[1]
\WHILE{$l=1:|L|$}
\STATE calculate $n_{l}$ based on (3)-(5).
\ENDWHILE
\IF{Network pruning}
\WHILE{$l=1:|L|$}
\STATE Pruning filters in layer $l$ based on a certain criterion.
\ENDWHILE
\ELSE
\STATE Generating the structure of student network $W$.
\ENDIF
\STATE \textbf{return} $W$ 

\end{algorithmic}
\end{algorithm}

\section{Experiments}
We reproduce tremendous methods on filter pruning and knowledge distilling to performance the effectiveness of the purposed method on the application these mothods. 
\subsection{Experimental Setup}
We train all the networks normally from scratch as baselines. We conduct compression for representative networks to evaluate the applicability of the proposed method to existing network pruning and model distillation methods. The representative networks are VGGNet\cite{simonyan2014very}, ResNet\cite{He_Zhang_Ren_Sun_2016} respectively. We report the performance on MNIST, CIFAR10 and ImageNet datasets, and compare to other automatic methods. The CIFAR10 dataset contains 50,000 training imges and 10,000 testing images with resolution $32\times32$, which are categorized into 10 classes. ImageNet dataset contains 1.28 million training images and 50,000 validation images of 1000 classes.\\
\vpara{Baselines.}
To demonstrate the effectives of the proposed method, we reproduce the classical filter pruning methods \cite{Li_Kadav_Durdanovic_Samet_Graf_2017a} based on $l$1-norm(l1) criteria and knowledge distilling methods  KD\cite{Hinton_Vinyals_Dean_2015} \& FT\cite{kim2018paraphrasing}, and then compare the results with an usual achitecture setting method, $i.e.$, compressing the filters equally in each layer. We also demonstrate the performance of our method on different compression works and reproduce other automatic methods to compare in a same compression criteria, including dynic pruning method \cite{he2018soft} based on $l$2-norm(l2) Geometric Median(GM) criteria \cite{he2019filter}, global pruning method based on batch normalization(BN) criteria \cite{liu2017learning}. For \cite{liu2017learning} introduces the scaling factors in BN Layers, we train a new network from scratch as the baseline for comparasion for 3AS and global pruning, the scaling factor is same as \cite{liu2017learning}, $\lambda=10^{-4}$. We do not compare the proposed method with heuristic-based and reinforcement learning-based compression methods, because they are difficult to design a feasible search granularity. These methods unify both architecture search process and pruning or distilling process where there is not a common criteria which can guarantee the fairness of the comparison results. To reduce the randomness of the experiments, we did every experiment three times and recorded the median as the final result.\\  
\vpara{Implementation Details.}We use Parameter and FLOPs (floating point operations) to evaluate model size and computational requirement respectively. We also demonstrated the effect of 3AS in different compression demands through filter compression ratio. We use Pytorch to implement the proposed 3AS. All the networks use the Stochastic Gradient Descent algorithm (SGD) with momentum 0.9, the weight decay is set to 1e-4. On CIFAR10, we train them with batch size of 64 and 160 epochs for baseline. The initial learning rate is set to 0.1, and it is divided by 10 when it arrives at 50$\%$ and 75$\%$ of the total number of the training epochs. On ImageNet, the batch size is set to 256,and the training epochs is set to 90, which is divided by every 30 epochs. We retrain the network for 40 epochs after pruning for fine-tuning on CIFAR10 and 30 epochs on ImageNet. And the learning rate is divided by 10 at 50$\%$ and 75$\%$ of the total number of the training epochs on CIFAR10 and divided by every 10 epochs on ImageNet.
\subsection{Compression on exsiting methods}
\vpara{Analysis on Adaptiveness.} We demonstrate the filter compression performance with VGG16 and VGG19 as the backbones on CIFAR10 \textit{w.r.t.} different compressing ratio. We compare the proposed method with equal proportion compression method(ECP) based on l1-norm \cite{Li_Kadav_Durdanovic_Samet_Graf_2017a}, KD\cite{Hinton_Vinyals_Dean_2015}, and FT\cite{kim2018paraphrasing}.
\begin{figure}
\centering
\subfloat[l1 compression]{
  \includegraphics[width=.5\linewidth]{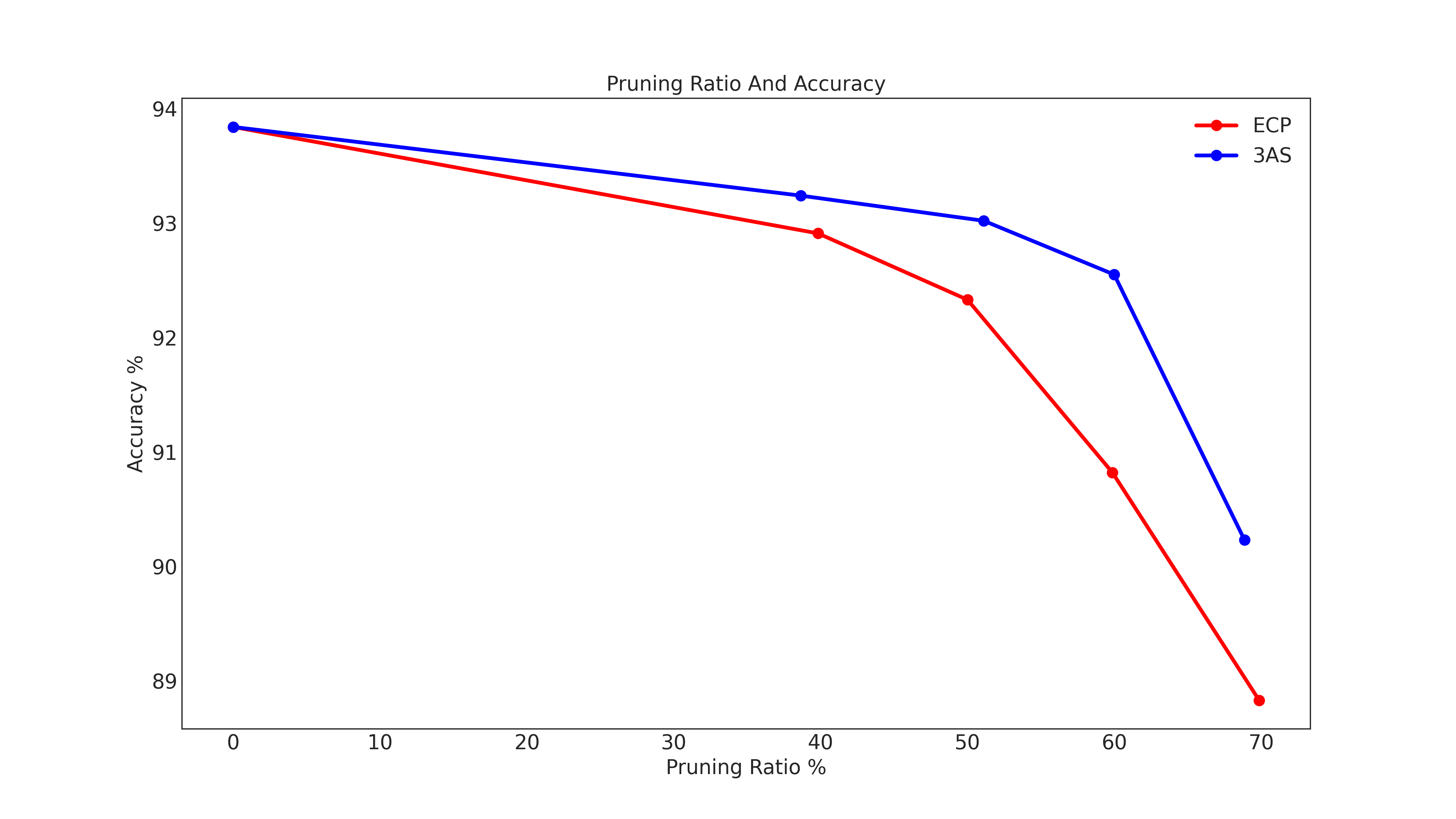}
}
\subfloat[KD compression]{
  \includegraphics[width=.5\linewidth]{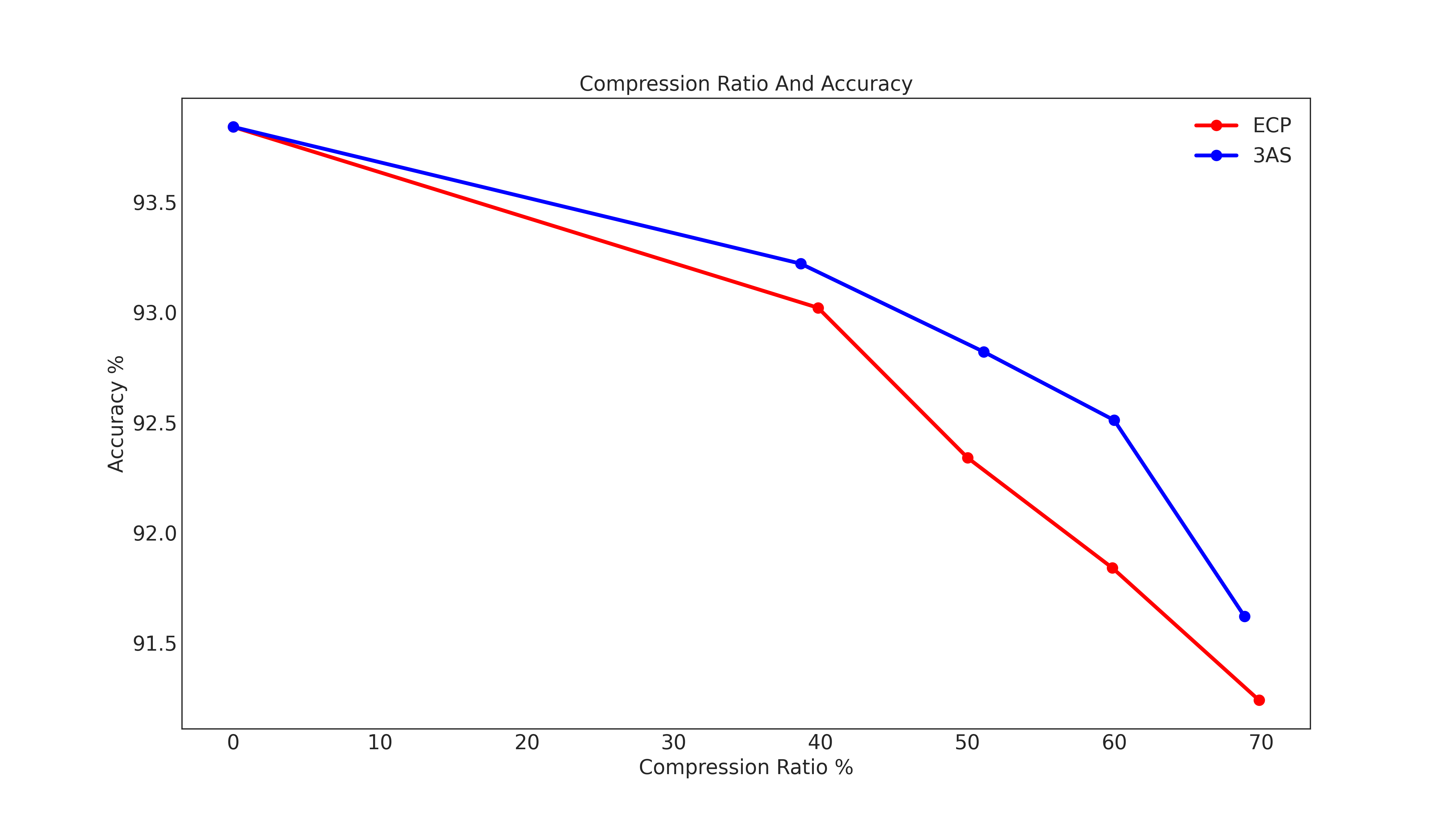}
}
\caption{Performance of 3AS and ECP with respect to different overall compression ratio on pruning and distilling method}
\end{figure}
For figure4(a) and figure4(b), it's obvious that the proposed method gets a better performance than the equal proportion compression method on both filter pruning method and knowledge distilling method. Remarkably, the performance gap between 3AS and the baseline method becomes significantly larger when the compression rate is larger, which validates that 3AS can identify the minimum number of filters that preserve more model capacity per layer. In other words, 3AS prefers to remove filters of over-paramterized layers, resulting in a more effective layer configuration under the same overall compression rate.\\
\vpara{Performance with different compression methods.}
We compare the proposed method with the equal proportion compression method on~\cite{Hinton_Vinyals_Dean_2015,Li_Kadav_Durdanovic_Samet_Graf_2017a,romero2014fitnets} and automatic method on \cite{he2018soft} and\cite{liu2017learning}.
\begin{table}
\centering
\scalebox{0.62}{\begin{tabular}{lrrr}
\toprule
Method & accuracy(\%) & flop pruned(\%) & param pruned(\%) \\ 
\midrule
Baseline(VGG16) & 93.84 & - & - \\
l1—ECP & 92.33 & 74.81 & 74.97 \\
l1—3AS & 92.53 & 75.72 & 87.09 \\
l2—dynic pruning\cite{he2018soft} & 92.31 & 74.81 & 74.97 \\
l2—3AS& 92.35 & 75.72 & 87.09 \\
GM—dynic pruning\cite{he2019filter} & 92.85 & 74.81 & 74.97 \\
GM—3AS& 92.94 & 75.72 & 87.09 \\
\midrule
Baseline(VGG19) & 93.83 & - & - \\
BN—global pruning\cite{liu2017learning} & 93.22 & 41.86 & 84.38 \\
BN—3AS & 93.39 & 54.74 & 84.38 \\
\bottomrule
\end{tabular}
}
\caption{Comparison with 3AS and architecture determination baselines on network pruning with respect to different pruning methods and backbones.}
\label{tab:tab1}
\end{table}
In table1, we show the performance of the proposed method on network pruning. For l1-norm, 3AS outperforms ECP (92.53\% vs.92.33\% ) with more parameter compression (87.09\% vs.74.97\%) and more speed up ratio (75.72\% vs.74.81\%). Also for automatic methods, 3AS has surpassed \cite{he2018soft} with a higher accuracy and better compression with a similar speed up ratio, also, on the same dynic pruning method \cite{he2019filter}, 3AS has a better performance for the better pruning criterion. And On VGG19, the global pruning method is just adopted for networks of a specific design, but the proposed method can also achieve a better performance on a desiged network than \cite{liu2017learning} with 0.83\% accuracy 
promotion in better compression.
\begin{table}
\centering
\scalebox{0.65}{
\begin{tabular}{lrrr}
\toprule
Method & accuracy(\%) & flop pruned(\%) & param pruned(\%)\\
\midrule
Baseline(VGG16) & 93.84 & - & - \\
KD-ECP & 92.34 & 74.81 & 74.97 \\
KD-3AS & 92.47 & 75.72 & 87.09 \\
FT-EPC & 92.10 & 74.81 & 74.97 \\
FT-3AS & 92.94 & 75.72 & 74.97 \\
\bottomrule
\end{tabular}}
\caption{Comparison with 3AS and architecture determination baselines on knowledge distillation with respect to different distillation methods.}
\label{tab:tab2}
\end{table}
In table2, the proposed method aims at provide an procipate architecture as the student network. For KD, compared with ECP, 3AS yields a less accuracy drop (92.47\% vs. 92.34\%) with more compression (74.97\% vs. 87.09\%) and a better speedup ratio. For FT, 3AS achieves 0.47\% accuracy improvement compared with 3AS in KD in the same compression rate. Not surprisingly, PCA get a better perfomance compared with ECP on accuray (92.94\% vs. 92.10\%), paramter compression and speedup ratio. These results demonstrate the superiority of 3AS on exsiting compressing method.
\begin{table}
\centering
\scalebox{0.65}{
\begin{tabular}{lrrrr}
\toprule
Method & Top-1 accuracy(\%) & param pruned(\%) & flop pruned(\%) \\ 
\midrule
l1-handcrafted\cite{Li_Kadav_Durdanovic_Samet_Graf_2017a} & 72.48  & 7.50 & 7.20 \\
l1-3AS & 72.61 & 7.67 & 7.88 \\
\bottomrule
\end{tabular}}
\caption{Comparison results on ImageNet}
\label{tab:tab3}
\end{table}
We also compared our methods applicated on l1-norm with hand-craft architecture designed in l1-norm\cite{Li_Kadav_Durdanovic_Samet_Graf_2017a}. As it is shown in table3, the proposed method has surpassed the best results in \cite{Li_Kadav_Durdanovic_Samet_Graf_2017a}, with high accuracy(72.61\% vs. 72.48\%), under a similar compression rate. Our method can not only find an approciate architecture but also efficient and computation-friendly compared with hand-crafted architecture.
\subsection{Ablation Study}
\begin{figure}
\centering
\subfloat[L1-norm on CIFAR10 ]{\includegraphics[scale=0.03]{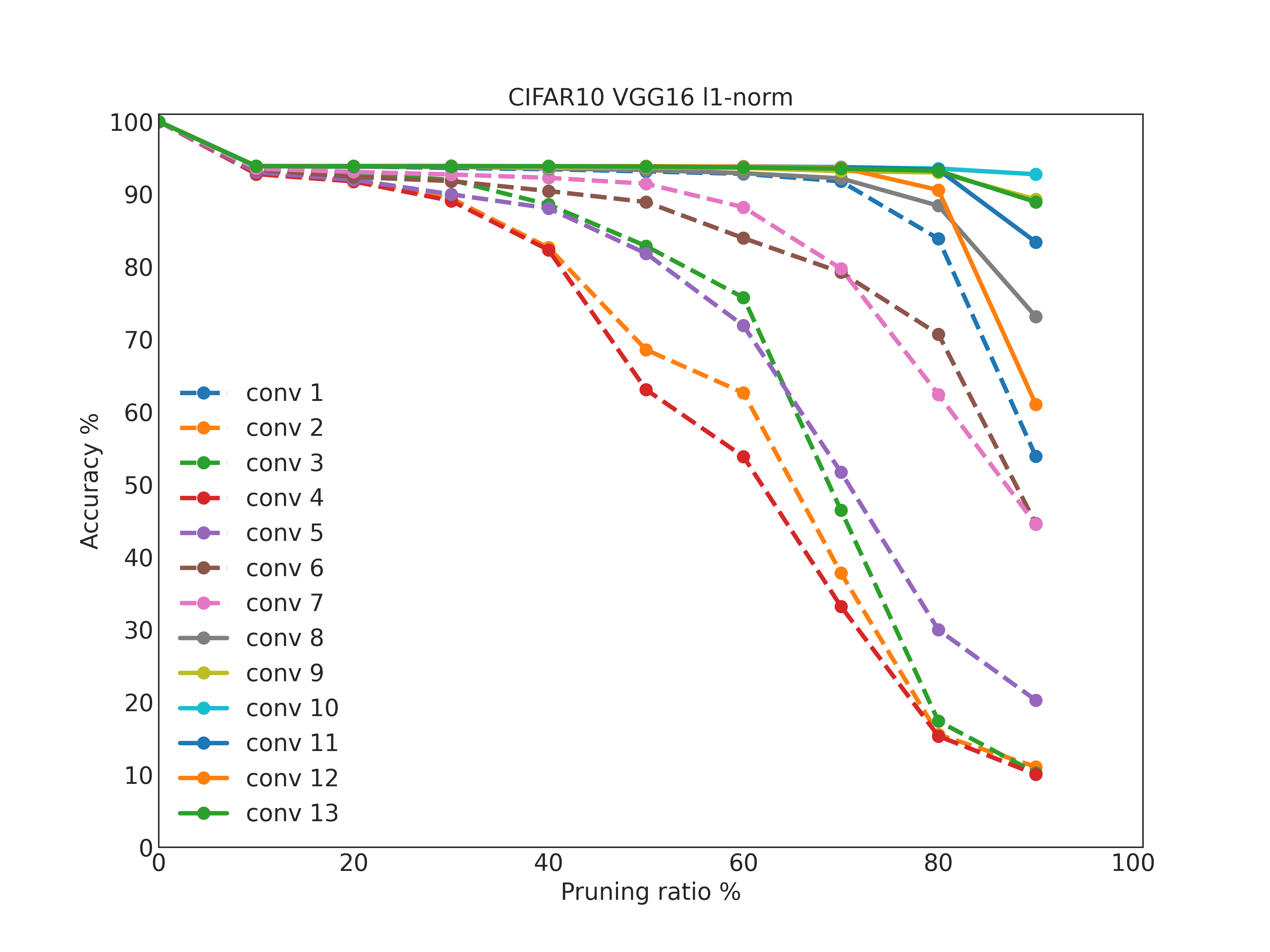}}

\subfloat[3AS on CIFAR10]{\includegraphics[scale=0.03]{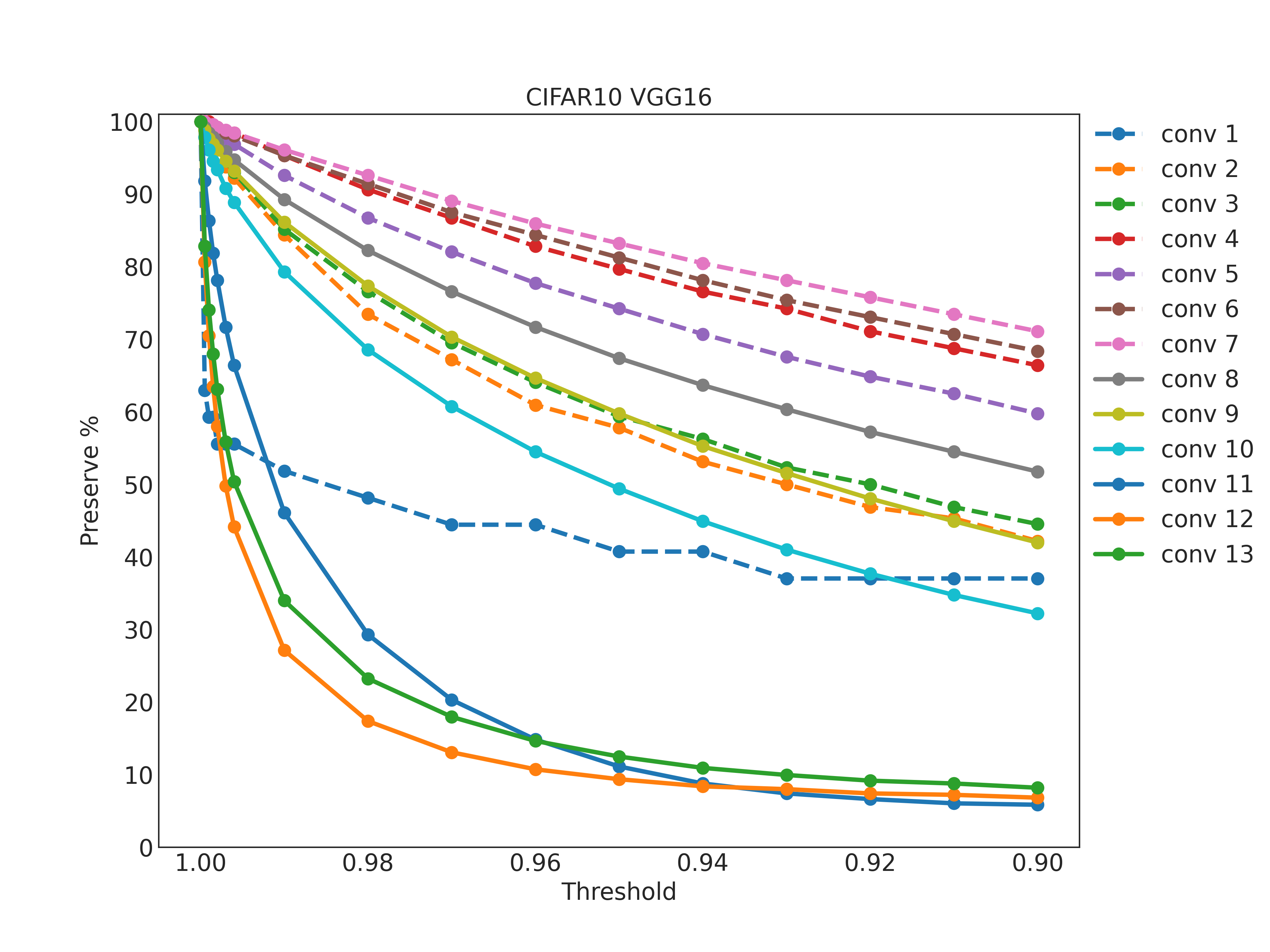}}
\subfloat[3AS on ImageNet]{\includegraphics[scale=0.03]{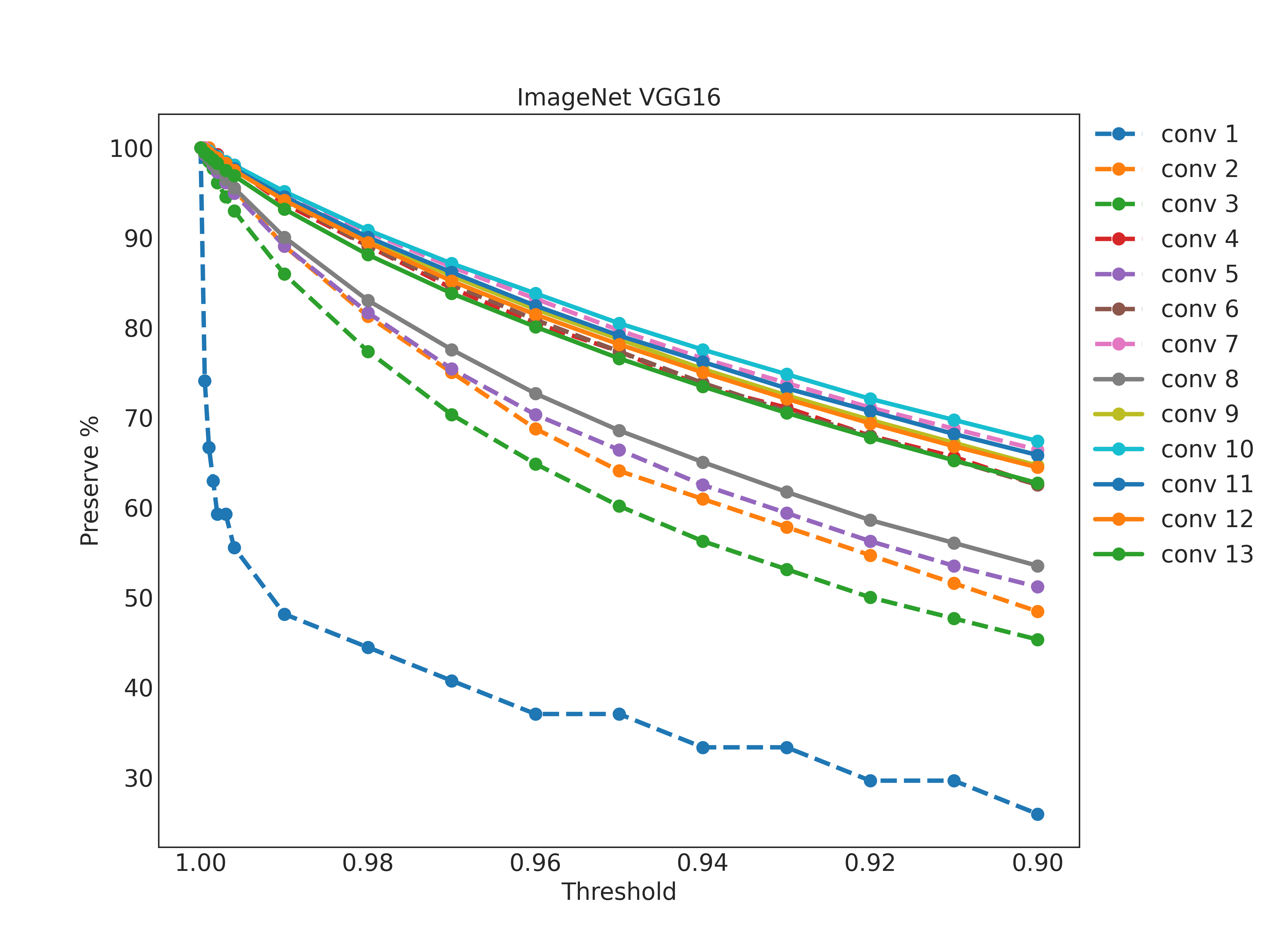}}
  \caption{(a) Filters preserving results of different conv layers based on l1-norm \textit{w.r.t.} different overall compression rates on CIFAR10.(b) Filters preserving results based on 3AS on CIFAR10(c) Filters preserving results based on 3AS on ImageNet.}
\end{figure}
To further demonstrate the effectiveness of 3AS, we analyzed the preserve ratio of filters at each layer calculated by 3AS with respect to different thresholds. \cite{Li_Kadav_Durdanovic_Samet_Graf_2017a} identify the sensitivity of all layers in CNN and analyze it in each layer through extensive experiments, we reproduce the results in figure5. By comparing (a) and (b), we can clearly find that under a certain thershold, PCA tend to preserve more filters in the former part of layers than the latter part of layers, which is consistent to (a), where the former layers is more sensitive than the latter layers. For example, 3AS tend to reduce the largest number of the last three layers(conv11,conv12,conv13), which are also the most insensitive layers in (a). 
Comparing (b) and (c), VGG is trained in different datasets, we can find that under the same threshod, 3AS tends to preserve more filters in ImageNet dataset than that in CIFAR10. Since ImageNet is larger than CIFAR10 where we might expct the filters is less redundent as those trained on CIFAR10. In other words, the redunbancy of filters in a layer correlates with its sensivity. 3AS can adaptively dertermine the compression architecture where the compression rate in each layer corresponds to the senstivities found by \cite{Li_Kadav_Durdanovic_Samet_Graf_2017a}. Note that 3AS obtains the layer-wise compression rates without tremendous experimentations, further demonstrating the merits of 3AS.

\section{Conclusion and Future Work} \label{sec:conclusion}
In this paper, we introduce a method, namely 3AS, that automates the process on network pruning and knowledge distillation. 3AS aims to find an appropriate compression architecture under a controllable threshold efficiently, where the time complexity corresponds to a one-step orthogonal linear transformation.
It is worth to mention that we find 3AS can analyze the sensitivity of each layer of CNN in an adaptive way. Thanks to this, 3AS can achieve a better performance compared with baseline methods and other automatic methods. 3AS can adaptively refine the layer configuration with respect the changable overall compression ratio. In the future, we aim to extend our idea to pruning technique with rigorous theretical information guaranee so as to achieve a more effective model compression with the cooperation of 3AS.
\newpage


\bibliographystyle{named}
\bibliography{sections/9.citations}

\end{document}